# Question answering using deep learning in low resource Indian language Marathi

Dhiraj Amin[1], Sharvari Govilkar[1], Sagar Kulkarni[1]

[1]Department of Computer Engineering, Pillai College of Engineering, Navi Mumbai, Maharashtra, India

*Abstract:* Precise answers are extracted from a text for a given input question in a question answering system. Marathi question answering system is created in recent studies by using ontology, rule base and machine learning based approaches. Recently transformer models and transfer learning approaches are used to solve question answering challenges. In this paper we investigate different transformer models for creating a reading comprehension-based Marathi question answering system. We have experimented on different pretrained Marathi language multilingual and monolingual models like Multilingual Representations for Indian Languages (MuRIL), MahaBERT, Indic Bidirectional Encoder Representations from Transformers (IndicBERT) and fine-tuned it on a Marathi reading comprehension-based data set. We got the best accuracy in a MuRIL multilingual model with an EM score of 0.64 and F1 score of 0.74 by fine tuning the model on the Marathi dataset.

## 1. Introduction

Natural Language Processing (NLP) is widely used to solve question answering tasks. Lots of work is done in the English language to solve reading comprehension-based question answering tasks where answers are extracted from relevant passages for given input questions provided in natural language. Marathi is the official language of Maharashtra state in India with more than 90 million native speakers, it is also in the top ten rank for most native speakers per language and the first language for of many people of Mumbai and Pune. Devanagari script is most widely used for writing Marathi content. Compared to English and Hindi language still a lot of work needs to be done to develop golden standard resources and dataset for Marathi language.

Marathi being a low resource and morphologically rich language compared to English language and does not have vast amounts of datasets like English for question answering tasks. Lots of English language datasets are available like Stanford Question Answering Dataset (SQuAD1.1) [1] and Stanford Question Answering Dataset (SQuAD2.0) [2] which led to speedy progress for development of models to solve English question answering tasks. Marathi language-based question answering work is less because of non-availability of such datasets in Marathi language. Apart from the English language, lots of work is done to develop native reading comprehension-based datasets for Chinese, Persian [6], Korean and other languages. For low resource languages like Marathi, language specific dataset can be developed by translating existing English language dataset into Marathi language. The quality of such a dataset depends on the accuracy of translation service. Also, existing English language dataset are built over content which are more related to English language speaking countries, such dataset does not have much information about Indian related content. Another approach is to build a cross lingual question answering system which can provide answers to Marathi questions without any requirement of training in Marathi Language.

Based on the research we introduce the MrSQuAD dataset, a novel extractive reading comprehension dataset specifically tailored for developing deep learning-based question answering systems in the Marathi language. The MrSQuAD dataset addresses the scarcity of Marathi language resources by providing a comprehensive and diverse collection of passages and corresponding questions. We investigate and address the challenges associated with the efficient translation of English text into Marathi, a low-resource Indian language, to create high-quality datasets. We provide and efficient process to translate English text into Marathi. This contribution aims to facilitate the creation of datasets in low-resource languages, such as Marathi, by providing insights and practical guidance. We conduct an empirical analysis of various state-of-the-art monolingual and multilingual transformer-based language models in the context of Marathi reading comprehension. We evaluate the accuracy and performance of these models on the task of extractive question answering by fine-tuning them using the Marathi reading comprehension dataset. This contribution presents a comparative assessment of different models and highlights their strengths and weaknesses when applied to Marathi question answering tasks.

## 2. Related Literatures

Recently transformer-based models are widely used to solve Natural Language Processing tasks like question answering, machine translation, name entity recognition, text summarization and part of speech tagging. Many transformer models are developed for the English language like Bidirectional Encoder Representations from Transformers (BERT) [3], Generative Pre-Trained Transformer (GPT) [19], Text-to-Text Transfer Transformer (T5) [21] and large multilingual models are also available like multilingual BERT (mBERT) [4], Multilingual Representations for Indian Languages (MuRIL) [5] and multilingual Text-to-Text Transfer Transformer (mT5) [20]. Large multilingual models are created which are very much useful to perform different tasks on low resource languages. Many languages like Hindi and Marathi share common structure and script, multilingual models built using similar languages can provide better resources for low resource language in the pair or group of languages.

Many Marathi language-based transformer models like MahaBERT [25] were also released which can be used to solve challenging Marathi language NLP problems. Previous works show that a language model trained on one language outperforms a model trained on more than one language. BERT [3] is a bidirectional attention-based transformer model trained on unlabeled corpus which can be fine-tuned on various downstream tasks like extraction question answering. The BERT architecture is used to create different forms of BERT like Robustly Optimized BERT Pre-training Approach (RoBERTa) [8], A Lite BERT (ALBERT) [9] and Distillable Bidirectional Encoder Representations from Transformers (distilledBERT) [7] and is trained in monolingual and multilingual modes.

Very little work is done to extract answers from Marathi language questions. English ontology was used by Chaware et al. [10] to answer Marathi questions. Onto terms extracted from English translated questions were used to extract answers from ontology. Govilkar et al. [11] developed semantic question answering using ontology created using domain experts. Subject, object, and verb were extracted from questions which helped in generating onto triples which finally extracted exact answers from ontology. S. Kamble et al. [12] scrapped questions and manually translated them into Marathi to create a Marathi Question Classifier using 1000 questions. The classifier was created using a direct and translation approach and got the best accuracy using a translation approach with 73.5% for coarse-grained class and 47.5% for fine-grained class. Darshan Navalakha et al. [13] created a chatbot system in Marathi language where the system automatically replied to user questions by extracting answers from pdf and images using Optical Character Recognition (OCR). Bharat Shelke et al. [14] created a question database for Marathi language. The questions were extracted from the Balbharti book for the 2nd to 4th standard classes. Total 901 questions were extracted from 50 different question types. Around 307 questions were of question type "what" as the questions were extracted from the student textbook. Aarushi Phade et al. [15] built Marathi language question answering systems by fine tuning Multilingual BERT model on a custom dataset of 1500 questions over Wikipedia and new articles and achieved F1-score of 56.7% and Bert-score of 69.08%. Anuj Gopal et al. [16] developed an automatic question generation model for Marathi using GPT [19] and T5 [21] by translating Marathi paragraphs in English language and post-processing was performed to remove any structural inaccuracy using Part-of-Speech (POS) and Named Entity Recognition (NER).

## 3. Research Methodology

The development of a Marathi question answering system starts with the creation of a specialized dataset tailored specifically for Marathi question answering. This dataset comprises a wide range of Marathi questions and their corresponding answers, covering diverse domains and topics. Next, different language transformer models are explored and analyzed to determine their suitability for the task. These models are then fine-tuned using the Marathi question answering dataset, enabling them to learn the intricacies of the Marathi language and optimize their performance. Throughout the process, dataset creation, model selection, fine-tuning, and evaluation are conducted to ensure the development of an accurate and efficient system capable of effectively addressing Marathi questions.

### 3.1. Marathi question answering dataset

The dataset for Marathi Question Answering is not available in the form of reading comprehension dataset which consists of context, questions, and answers. Existing datasets are in the form of question and answers only [10] [11]. We have created a Marathi dataset for reading comprehension tasks by using SQuAD1.1 [1] dataset. A single line consists of a context which is a passage of a few sentences, question related to the passage, answer for the question asked over the passage and answer start ID indicating start index or position of answer present in the passage. The SQuAD1.1 [1] dataset in the validation set was having more than one correct answer for a given question, to reduce complexity we have automatically selected the answer which occurs a greater number of times compared to other answers in the list of answers for a given question.

We have manually filtered out unwanted context and corresponding question and answers from the existing English SQuAD1.1 dataset. The English SQuAD1.1 dataset consists of many contexts which are not suitable for translation to Marathi Language. In the SQuAD1.1 dataset many non-English words are also present which are not needed in the contexts, questions, and answers. Also, in SQuAD1.1 dataset many contexts are related to information of characters and phonetics of different languages written in English which is also filtered out from the dataset. After removing unwanted contents from existing dataset, the dataset is further translated to Marathi language using translation service. Some of the translated sentences have words which are not translated to Devanagari script. Script validation is performed to identify the words in Latin script which are not translated correctly into Marathi language. Such words are further transliterated into Devanagari script using AI4Bharat transliteration [26] library to get the most accurate Marathi text. Most of the translation services fail to translate English digits into Devanagari form during the translation process. In the dataset there are many numbers in terms of quantity, price, or dates. After translation we are processing the dataset by automatically detecting such digits in Marathi translated text and translating it into Devanagari script to improve the quality of translated dataset. Figure 1 demonstrates the translation and transliteration of English sentences with characters and numbers into Marathi sentences.

During translation of text from English to Marathi language sometimes the translated word in question or context does not match with translated answers. We have also filtered out such instances from the dataset as it will lead to incorrect question answer pairs from the passage. Further "." was observed at the ending of many translated answers and it was also removed from the answers as we needed to recalculate start ID. If "." is not removed from the answer span, then while fine tuning language models for question answering many question answer pairs were not considered for feature extraction which leads to shortened dataset. The start ID present in the SQuAD1.1 English dataset cannot be used for translation answers as after translation the position of answer in the passage will vary. The answer start ID is again recalculated for translated answers. This recalculation can sometimes mark answers at incorrect positions in the context. Also, it was observed that for many pairs of question and answer the start ID was not possible to calculate because translations of the same word or phrase were different in question and answer. This led to reduced size of dataset of both train and test splits.

The final translated dataset after filtration of unwanted pairs consists of 47065, 5832 questions, answer and passage pairs as train and test respectively. In the train part of the dataset 17337 unique contexts, 46960 unique questions and 30162 unique answers are present. In the test part of the dataset 1922 unique contexts, 5805 unique questions and 4409 unique answers are present.

**Figure 1.** Translation and transliteration of English text into Marathi text

The structure of a single context and its question answers of the Marathi dataset is shown in Figure 2. More than one question is present for a passage and most of the questions are factoid and short answers are provided for such questions. All the questions in the dataset are answerable questions same as SQuAD1.1 format.

**Figure 2.** Marathi question answer dataset structure in SQuAD format

### 3.2. Marathi pre-trained language models

Recently few multilingual and monolingual language models have been released supporting Marathi languages. Most of the language models are developed using the fill mask method which does not require any labeled data. In the fill mask method few of the words in the sentence are masked out and the trained model is expected to predict the correct masked word in the given sentence. Such models have statistical information of words, sequence of words and occurrence of words for the language. These models can be further trained to solve question answering tasks. By fine tuning the existing multilingual and monolingual model, simple question answering systems can be created using smaller annotated question answering dataset of Marathi language.

There are four multilingual language models supporting Marathi language. They are DistilBERT multilingual [7], MuRIL [5], Cross-lingual Language Model – RoBERTa (XLM-RoBERTa) [17], Indo-Aryan Cross-lingual Language Model - RoBERTa (Indo-Aryan-XLM-R) and BERT Multilingual [2]. BERT multilingual [2] is trained on 104 languages using the Wikipedia dataset on masked language modeling and next sentence prediction approach. Marathi was one the language used for training mBert transformer models. We have used bert-base-multilingual-cased to mBert conduct our experiments. The model is case sensitive in nature and performance better than case insensitive mBert model. Cross-lingual Language Model - RoBERTa (XLM-RoBERTa) [17] is a multilingual version of the RoBERTa model which is trained on 100 languages including Marathi on CommonCrawl data of size 2.5TB. The RoBERTa transformer model was developed using a masked language modeling approach. We have used the xlm-roberta-base model to conduct our experiments on a Marathi annotated dataset. DistilmBERT [7] is a distilled version of Bert multilingual model which is twice faster and much smaller in size compared to mBert. The model does not outperform mBert model but the accuracy of the DistilmBERT is not much less compared to mBert transformer model. We have used a distilbert-base-multilingual-cased model to conduct our experiments. IndoAryanXLMR is fine-tuned on XLM-RoBERTa [17]. It is trained using Open Super-large Crawled Aggregated coRpus (OSCAR) [23] monolingual corpus consisting of Marathi, Hindi, Gujarati, and Bengali languages. The model was trained on a masked language modeling approach after balancing the dataset. We have used the Indo-Aryan-XLM-R-Base model to conduct experiments. MuRIL [5] is pre-trained in 17 Indian languages including English. The data for training is extracted from Wikipedia, Common Crawl, PMINDIA and parallel translated and transliterated data is used. It is a BERT base architecture trained on masked language modeling tasks with 512 as max sequence length. It outperformed multilingual BERT on many downstream tasks. We have used muril-based-cased for our experiment. Indic Bidirectional Encoder Representations from Transformers (IndicBERT) [18] is a multilingual ALBERT based model pre-trained on 12 Indian languages including Marathi. It trained on 452.8 million sentences of the monolingual corpus of 12 languages combined. It has fewer parameters compared to mBERT and XLM-R but it works accurately or sometimes better than these models on downstream tasks. We have used the base model of ai4bharat/indic-bert. DevBERT [24], DevRoBERTa [24], DevAlBERT [24] are forms of the Devnagri BERT model. They are multilingual models trained over the monolingual corpus of Hindi and Marathi languages as both the languages share the same Devanagari script. All the models are trained over 2.5 billion Devanagari tokens with the same hyperparameters and with 2e-5 as learning rates. The models tend to perform better compared to existing multilingual models. A version of DevBERT model is also created where the model is trained from scratch using a custom trained tokenizer. We have used all the four versions for our experiment.

There are many Marathi monolingual models released recently. Majority of them are released by L3Cube Pune. They are based on a variation of BERT and trained on Marathi corpus. MahaBERT [25] model, MahaAlBERT [25] model and MahaRoBERTa [25] model is fine-tuned on a Marathi monolingual corpus of 57.2 million sentences, 752 million tokens and are based on multilingual BERT, IndicBERT based monolingual AlBERT and multilingual RoBERTa based architectures. All the models were trained using a masked language objective for Marathi language only. All the models were trained for 2 epochs on the same hyperparameters with 2e-5 learning rate with a batch size of 64. The models were evaluated on downstream tasks like text classification, NER and outperformed the existing multilingual models. A version of the MahaBERT model is also created using custom trained tokenizers and trained from scratch instead of fine tuning on existing multilingual models. We have used marathi-bert-v2, marathi-roberta and marathi-albert-v2 for our experiments. Marathi DistilBERT is trained from scratch using DistilBERT architecture which is a lightweight distilled version of BERT model. The model was trained over Oscar Corpus, Marathi Newspapers, Marathi story books and articles totaling around 11.2 million sentences with learning rate 1e-4 and mask probability of 15%. We have used marathi-distilbert for our experiments.

## 4. Results and Discussions

Multilingual and monolingual models are evaluated over Marathi reading comprehension-based dataset. The results are described in Table 1 and Table 2 for multilingual and monolingual models fine-tuned on Marathi Question answering dataset. We observed that best monolingual model trained only on Marathi language and multilingual model trained on different languages including Marathi have almost similar results for finding out the answers for the given natural language questions asked over the passage as a form of extractive question answering systems.

### 4.1. Experimental-Setup

All the language models were fine-tuned on a Marathi dataset having 47065 questions and passages pairs and tested on 5832 questions and passages pairs. The output of the fine-tuned models is probable starting and ending index of the predicted answers for all the models and all the models were trained for 2 epochs with batch size set to 15. All the models were trained using GPU with sequence length set to 512. Approximately it takes on average one and half to two and half hours to train for 2 epochs for different models.

### 4.2. Evaluation Metrics

For a given input question and passage the model predicts the most correct answer in terms of starting and ending index of sequence in the passage. Many of the predicted sequences will have words which are not part of ground truth answers or it may be semantically like the words present in the ground truth answers. Exact Match (EM) score, F1 score and BERT score [22] are few of the evaluation metrics which are used to evaluate extractive question answering models.

For reference answer: मराठी प्रश्न उत्तर शिकणे (learning Marathi question answer - in English) and predicted answer: मराठी प्रश्न उत्तर शिकणे (learning Marathi question answer - in English) the EM score is 1.0, F1 Score is 1.0 and BERTScore is 1.0 as all the words are same in correct and predicted answer. For reference Answer: मराठी प्रश्न उत्तर शिकणे (learning Marathi question answer - in English) and predicted answer: मराठी शिकणे (learning Marathi - in English) the EM score is 0.0, F1 Score is 0.67 and BERTScore is 0.88 here as only two words match the scores are low compared to previous example. For reference answer: मराठी प्रश्न उत्तर शिकणे (learning Marathi question answer - in English) and predicted answer: मराठी प्रश्न समाधान शिकणे (learning Marathi question solution - in English) the EM score is 0.0, F1 Score is 0.75 and BERTScore is 0.91. Here BERTScore is increased as उत्तर (answer in English) and समाधान (solution in English) are semantically related with each other.

### 4.3. Results

We fine-tuned xlm-roberta-base, muril-base-cased, Indo-Aryan-XLM-R-Base, bert-base-multilingual-cased, distilbert-base-multilingual-cased on Marathi Question answering training set with same hyper-parameters. The results from Table 1 shows that muril-base-cased models achieved the highest score for EM, BERT and F1. Indo-Aryan-XLM-R-Base achieved slightly higher results compared to xlm-roberta-base, which specifies that fine tuning existing models with language specific datasets gives better performance. We noticed BERT based models had better results compared to RoBERTa.

Table 1. Comparison of different multilingual models over Marathi QA dataset

| Model | EM Score | BERT Score | F1 Score |
|---|---|---|---|
| DistilBERT Multilingual | 0.50 | 0.88 | 0.60 |
| mBERT | 0.59 | 0.90 | 0.69 |
| XLM-RoBERTa | 0.42 | 0.90 | 0.58 |
| IndoAryanXLM-RoBERTa | 0.43 | 0.90 | 0.58 |
| MuRIL | 0.64 | 0.93 | 0.74 |

We fine-tuned different Marathi monolingual models like marathi-bert-v2, marathi-albert-v2, indic-bert, marathi-roberta, marathi-distilbert, hindi-marathi-dev-bert, hindi-marathi-dev-roberta, hindi-marathi-dev-albert, marathi-bert-scratch on Marathi Question answering training set with same hyper-parameters. The results from Table 2 shows marathi-bert-v2 and hindi-marathi-dev-bert were the best performing models compared to others.

Table 2. Comparison of different monolingual models over Marathi QA dataset

| Model | EM Score | BERT Score | F1 Score |
|---|---|---|---|
| MahaBERT | 0.63 | 0.93 | 0.73 |
| Indic BERT | 0.32 | 0.84 | 0.44 |
| MahaRoBERTa | 0.42 | 0.90 | 0.58 |
| MahaAlBERT | 0.37 | 0.87 | 0.50 |
| Marathi DistilBERT | 0.22 | 0.78 | 0.30 |
| DevBERT | 0.63 | 0.92 | 0.73 |
| DevRoBERTa | 0.43 | 0.90 | 0.58 |
| DevAlBERT | 0.37 | 0.87 | 0.50 |
| DevBERT-scratch | 0.27 | 0.80 | 0.35 |

The dataset on which monolingual and multilingual transformer model is trained also affects the accuracy of the system. Hindi and Marathi almost share the same grammatical structure which can improve performance of training language models with much larger datasets. The results of the best multilingual and best monolingual model are almost similar. We tried training the certain models for 4 epochs and different learning rates but there was no significant increase in the results. The results can be further improved if the Marathi question answering dataset is created from scratch using existing Marathi text instead of translations of English text to Marathi language.

## 5. Conclusion

Many researchers have gained interest in solving the natural language challenges over low resource languages like Marathi. SQuAD dataset was the baseline dataset which provided the researchers enough data to devise solutions to solve the question answering problem. Translation of existing English dataset to Marathi language can help as a starting path for building a deep learning-based question answering system for Marathi language. The release of the Marathi monolingual and multilingual language model has contributed to solving many natural language problems. Marathi extractive question answering system can be easily developed by fine tuning marathi-bert and muril language models. Development of a bigger dataset from scratch of Marathi question answers may provide better results compared to the translated dataset approach.